\documentclass[10pt,twocolumn,letterpaper]{article}

\usepackage{cvpr}
\usepackage{times}
\usepackage{epsfig}
\usepackage{graphicx}
\usepackage{amsmath}
\usepackage{amssymb}
\usepackage{gensymb}
 \usepackage{color}
\usepackage[breaklinks=true,bookmarks=false]{hyperref}

\cvprfinalcopy 

\def\cvprPaperID{****} 

\ifcvprfinal\fi
\begin{document}

\title{Multi-View Matching Network for 6D Pose Estimation}




\author{Daniel Mas Montserrat\textsuperscript{1}, 
\quad Jianhang Chen\textsuperscript{2}, 
\quad Qian Lin\textsuperscript{3}, 
\quad Jan P. Allebach\textsuperscript{2}, 
\quad Edward J. Delp\textsuperscript{1}\\
\textsuperscript{1}Video and Image Processing Laboratory (VIPER), Purdue University\\
\textsuperscript{2}Electronic Imaging Systems Laboratory (EISL), Purdue University\\
\quad \textsuperscript{3}HP Labs, HP Inc.\\
}


\maketitle

\begin{abstract}
Applications that interact with the real world such as augmented reality or robot manipulation require a good understanding of the location and pose of the surrounding objects. In this paper, we present a new approach to estimate the 6 Degree of Freedom (DoF) or 6D pose of objects from a single RGB image. Our approach can be paired with an object detection and segmentation method to estimate, refine and track the pose of the objects by matching the input image with rendered images.

\end{abstract}

\section{Introduction}
\label{sec:intro}
The main challenge when using augmented reality, robot manipulation or autonomous driving is to obtain a clear and detailed understanding of the geometry of the objects appearing in  image or video to properly interact with them. In order to understand how the objects in the scene are placed, the pose and location of each element need to be inferred. 6 Degree of Freedom (DoF) or 6D pose estimation techniques can be useful for such tasks. 6D pose estimation consist of inferring the 3D location coordinates and 3D rotation angles of objects in the real world from images, videos or depth information.


Many solutions have been presented for 6D pose estimation working with both RGB-D and RGB images. RGB-D cameras are not highly available (e.g. in smart-phones or laptops) and have many limitations in the depth range, resolution and frame rate making it difficult to detect small, thin or fast moving objects. On the other hand, methods that use RGB-only images can easily fail to correctly estimate the pose of objects with different lighting conditions, occlusions or objects that lack texture or distinctive visual features.

\begin{figure}[tb]

\begin{minipage}[b]{1.0\linewidth}
  \centering
    \centerline{\includegraphics[width=1.0\linewidth]{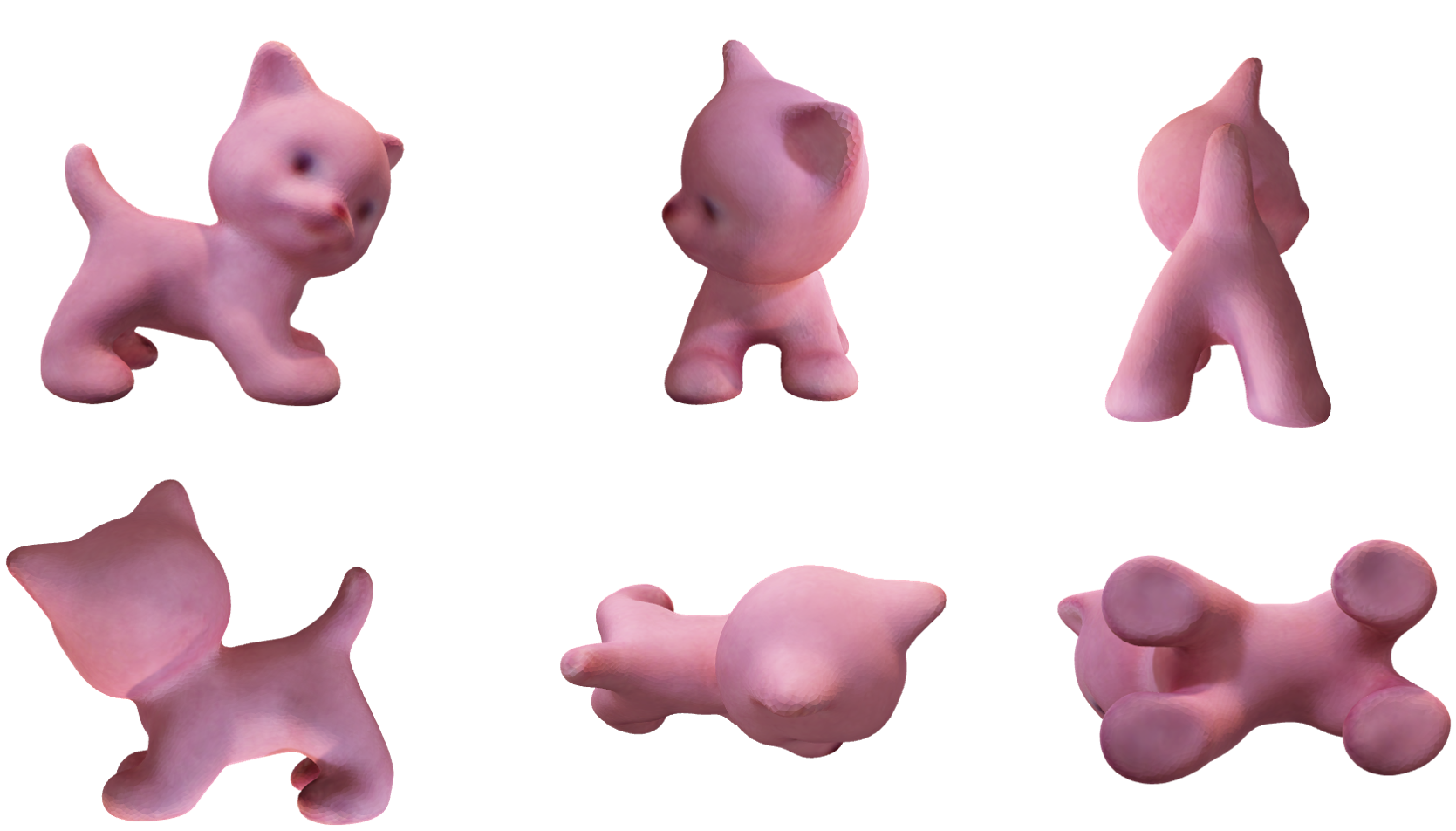}}
\end{minipage}
\caption{Six rendered views of an object.}
\label{fig:6views}
\end{figure}

We introduce a new set of neural networks, Multi-View Matching Network (MV-Net) and Single View Matching Network (SV-Net) for 6D pose estimation, refinement, and tracking. These networks are based on DeepIM \cite{deepim}, a neural network that performs pose refinement by estimating the pose difference between pairs of images.  MV-Net matches the input image with 6 RGB images containing the object of interest rendered from different views (Figure \ref{fig:6views}) to obtain an initial estimate of the pose. Then, SV-Net refines the pose estimate in an iterative manner. The same iterative refinement can be used to track the pose in a video sequence.

The main contributions of this paper are as follows. First, we present a new way to extend DeepIM, a pose refinement network, for pose estimation, refinement and tracking, removing the need for an external initial pose estimation method. Second, we show how these networks can be combined with Mask R-CNN to have a complete object detection and pose estimation pipeline. Finally, we evaluate and compare our method with previous pose estimation methods.


\section{Related Work}
\label{sec:related-work}

Traditional computer vision methods estimate the 6D pose of an object by matching visual features from the image to a 3D model of the object.  
The main drawback of this approach is that it fails with textureless objects since only a small number of visual features can be detected on the object. 
Many recent approaches based on deep learning treat pose estimation as a classification or regression task by extending object classification or detection networks \cite{posecnn, 3drcnn}. Although they can handle textureless objects, they are not highly accurate as the pose might be discretized into bins and the methods rely on learning the visual appearance of the objects at different poses. 
Another approach is to directly regress the 3-dimensional bounding box containing the object with a neural network as in \cite{BB8}.

Recently,  methods for pose refinement have been presented showing considerable improvements in accuracy \cite{deepim}. A common approach is to render an RGB image with an initial pose estimate and then match the rendered image with the input image to obtain a new pose estimate. 
One of these methods is DeepIM \cite{deepim}, the basis of our work. This method extends FlowNet \cite{flownet}, a neural network trained to estimate the optical flow between consecutive frames, to estimate the pose difference between images. 

Multi-view methods have been used in pose related problems. The method presented in \cite{MVCNN} uses a multi-view convolutional neural network for shape estimation. The method proposed in \cite{MVMC} uses a multi-view multi-class system for pose estimation.

\section{Proposed Method}
\label{sec:proposed-method}

\begin{figure}[htb]
\centering
\includegraphics[width=1.0\linewidth]{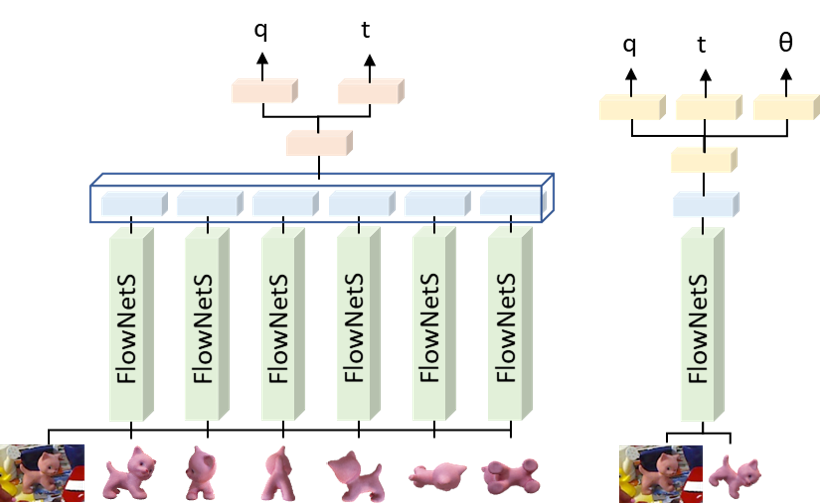}
\caption{Multi-View Matching Network (left) and Single-View Matching Network (right). }
\label{fig:network}
\end{figure}

Our proposed method has multiple stages. First, we use Mask R-CNN \cite{maskrcnn} to detect and segment the objects of interest in the input image. Next, we estimate the 6D pose of the objects with the Multi-View Matching Network. Finally, the pose estimates are refined with the Single View Matching Network. Figure \ref{fig:network} shows the MV-Net and SV-Net architectures.

\subsection{Object Detection And Segmentation}
\label{ssec:object-detection-and-segmentation}
Many object detection and segmentation methods are currently available \cite{yolov2, maskrcnn}. 
In our work, we use Mask R-CNN with ResNet-50 \cite{maskrcnn} as it provides state-of-the-art results on detecting and segmenting common objects.

We use the bounding boxes estimated by Mask R-CNN for a zoom-in operation to the regions where objects have been detected. The zoom-in operation consists of cropping and resizing to $640 \times 480$ the region inside the estimated bounding boxes. Previous to resizing, bounding boxes are expanded to have a 4:3 ratio in order to maintain the aspect ratio. The same zooming in operation is used for the estimated segmentation mask.

\subsection{Initial Pose Estimation}
Multi-View Matching Network (MV-Net) takes as input 6 rendered images with the detected object in 6 different poses and the zoomed target image to estimate the initial pose of the object.

MV-Net is composed of 6 parallel branches, each of them consisting on the first 10 convolutional layers of FlowNetSimple network followed by one fully-connected layer of dimension 256. The outputs of the fully-connected layers of each 6 branches are concatenated and followed by one fully-connected layer of dimension 512. Finally, two fully-connected output layers of dimension 4 and 3 are added to estimate the relative rotation and translation parameters respectively. The weights of every layer in the 6 branches are shared among them and with SV-Net.

Each branch has as input an 8 channel tensor composed by on the RGB zoomed target image, its segmentation mask (obtained from Mask R-CNN) and a rendered RGB image with its mask. This 8 channel input approach is the same as in \cite{deepim}. The rendered image used at each branch has a different pose. In our method, we select the 6 equidistant angles in the pitch and yaw dimensions, equivalent to the 6 views of the faces of a cube. Figure \ref{fig:6views} shows the rendered images of the 6 views of an object.

The estimated rotation and translation are represented as the relative pose of the target image and the object placed with its frontal face facing the camera (the input of the 1st branch of MV-Net). The initial location of the object is inferred from the center of the bounding box. 

The network learns the relative pose with the same representation as in \cite{deepim}, where the relative rotation is expressed with a quaternion and the relative translation is expressed with an untangled representation independent from the coordinate system of the object. Both quaternion and untangled translation parameters are normalized to have zero mean and unit variance.


\subsection{Pose Refinement}
Single View Matching Network (SV-Net) refines the MV-Net pose estimate. SV-Net consists of one single branch from MV-Net followed by one fully-connected layer of dimension 256. Then 3 fully-connected layers of dimension 4, 3 and 1 are used to estimate the relative rotation, translation and angle distance respectively. Note that without the angle distance estimation output, SV-Net is equivalent to DeepIM\cite{deepim}.

As in MV-Net, the input of the SV-Net consists of one 8 channel tensor composed by the target RGB+mask image, and an RGB+mask rendered image. The rendered image is obtained by rendering the object with the previously estimated pose. The refinement process starts with the initial pose estimate of MV-Net. Then, SV-Net estimates the pose difference between the rendered image and the zoomed input image. The pose difference is used to update the previous pose estimate. A new image is rendered with the updated pose and the refinement process is repeated until the estimated rotation angle, $\hat{\theta}$, between the two images is below a threshold ${T}_{ref}=2\degree$. If after 50 refinement iterations, the estimated rotation angle isn't below ${T}_{ref}$, the pose estimate with the lowest estimated rotation angle is selected. This policy to stop the refinement process differs from other methods (e.g. \cite{deepim}) where a fixed number of refinement iterations are used.

\subsection{Pose Tracking}
\label{ssec:pose-tracking}
As in DeepIM, SV-Net pose refinement can be easily extended for pose tracking. After the pose is estimated and refined in the first frame, the pose tracking is performed by estimating the pose difference between the current pose estimate and the next frame of the video. Then, the pose estimate is updated, and the process is repeated for the next frame. 

The angle distance estimate of the SV-Net can be used to assist the tracking process. If the angle distance estimate is higher than a threshold $T_{high}=25\degree$, we restart the tracking process by estimating a new initial pose with MV-Net. If the angle distance estimate is lower than a threshold $T_{low}=2\degree$, the pose estimate is not updated.

\subsection{Training Process}
\label{ssec:training-process}

Our networks are trained using the LINEMOD\cite{linemod} dataset and synthetic images  described in Section \ref{sec:image-generation}. These datasets include the 3D models of multiple household objects and images containing such objects. We use as ground truth the bounding box surrounding the objects of interest, their segmentation mask, and their pose represented with a rotation quaternion, and the 3D location values.

We fine-tune Mask R-CNN using stochastic gradient descent (SGD) with a learning rate of 0.005 for 10 epochs. We use pre-trained weights from the MS COCO dataset \cite{COCO}.

We train MV-Net and SV-Net using the loss functions in Equation \ref{eq:loss-m} and Equation \ref{eq:loss-s} respectively.

\begin{equation}
\label{eq:loss-m}
L_m(p, \hat{p}) = \frac{1}{N}(\|1-q^T\frac{\hat{q}}{\|\hat{q}\|}\|  + \|t-\hat{t}\| + \|1-\|\hat{q}\|\|)
\end{equation}

\begin{equation}
\label{eq:loss-s}
L_s(p, \hat{p}) = \|1-q^T\frac{\hat{q}}{\|\hat{q}\|}\|  + \|t-\hat{t}\| + \|1-\|\hat{q}\|\| + \|\theta-\hat{\theta}\|
\end{equation}

Where $p = [q|t]$ and $\hat{p} = [\hat{q}|\hat{t}]$ are the ground truth and estimated rotation quaternion and translation parameters respectively. 
$\theta$ and $\hat{\theta}$ are the ground truth and estimated angle distance respectively. 
$N$ is the number of views. In this work we use $N=6$.

We follow a simultaneous end-to-end training approach. The training process starts by estimating the initial pose with MV-Net.
Then, the initial estimate is used to render a new training sample for SV-Net. We perform a total of 3 refinement iterations during training time. 
In order to avoid training SV-Net with bad initial estimates, we generate a new sample with a small random rotation if the angle error of the initial estimate is higher than $25\degree$. 

We initialize the FlowNetS convolutional layers of MV-Net and SV-Net with pre-trained weights for optical flow estimation. The fully-connected layers are initialized with random weights as in \cite{deepim}. We train the networks using SGD with learning rate 0.001 during 48 epochs.

Our simultaneous end-to-end training approach has several advantages. First, we can easily apply weight sharing between MV-Net and SV-Net. Second, we present the training examples to the network in a similar same order as in testing time. Therefore, SV-Net learns to fix the mistakes produced by MV-Net. 


\subsection{Datasets}
\label{sec:image-generation}

\begin{figure*}[htbp]
\caption{Examples of photorealistic images (top row) and domain randomized images (bottom row).}
\centering
\includegraphics[width=\textwidth]{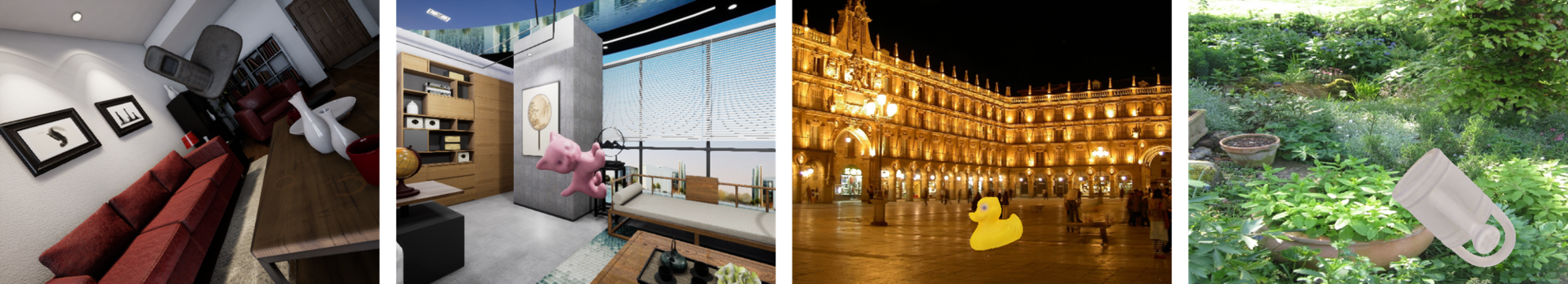}
\label{fig:photorealistic}
\end{figure*}

We use the LINEMOD (LM) dataset \cite{linemod} for training and testing. This dataset contains 15 objects but we only use 13 as in previous works \cite{deepim}. We use the same training and testing split as in \cite{deepim}. In addition, we use two different rendered datasets for training: A domain randomized dataset and a photorealistic dataset. Figure \ref{fig:photorealistic} shows some examples of both rendered datasets.

The domain randomized dataset is generated by rendering the objects in random poses and adding background images. The background images are randomly selected from Pascal VOC \cite{Everingham15} dataset. Additionally, we apply random blurring and color jittering to the image. The segmentation mask is dilated with a square kernel with a size randomly selected from 0 to 40 in order to mimic errors in the segmentation mask during inference. We render 5,000 training images for each object.

We the 3D models of LM dataset in virtual environments to generate photorealistic images with Unreal Engine 4 (UE4) \cite{unrealengine4}. This includes a total of 5,000 images per object.


\section{Experimental Results}
\label{sec:experimental-results}

We evaluate our method using the ($n\degree$, $n$ cm) accuracy metric with the LINEMOD testing set. This metric considers an estimate correct if the error is smaller than $n\degree$ and $n$ cm and incorrect otherwise. We evaluate our method for $n=2,5,10$. Table \ref{tab:results} presents our accuracy results.

\begin{table}[htb]
\begin{center}
\begin{tabular}{|l|c|c|c|}
    \hline
     & \multicolumn{3}{|c|}{($n\degree$, $n$ cm)} \\
    \cline{2-4}
     Method & (2, 2) & (5, 5) & (10, 10) \\

    \hline
    BB8 w/ ref. \cite{BB8} & - & 69.0\% & -\\
    PoseCNN \cite{posecnn} & - & 19.4\% & -\\
    PoseCNN+DeepIM \cite{deepim} & 39.0\% & 85.2\% & 97.9\% \\
    Ours & 24.3\% & 73.1\% & 97.5\%\\
\hline
\end{tabular}
\end{center}
\caption{Comparison with previous methods on LINEMOD dataset.}
\label{tab:results}
\end{table}

We can observe that our accuracy results are comparable with previous state-of-the-art RGB-only pose estimation methods such as PoseCNN+DeepIM \cite{deepim}. 

Our method shows that pose refinement methods \cite{deepim} can successfully be extended for initial pose estimation. Therefore, initial pose estimation networks such as PoseCNN \cite{posecnn} can be replaced by highly available object detection networks \cite{maskrcnn, yolov2} which can be trained with a large amount of training data.


\section{Conclusions}
\label{sec:conclusions}
In this paper, we present how a pose refinement network can be extended to perform both pose estimation and refinement. We introduce new research ideas to have a unique network to estimate, refine and track the pose of objects. We show how the network can automatically assist the refinement and tracking process.

{\small
\bibliographystyle{ieee_fullname}
\bibliography{all}
}

\end{document}